\documentclass[conference]{IEEEtran}
\IEEEoverridecommandlockouts
\usepackage{cite}
\usepackage{amsmath,amssymb,amsfonts}
\usepackage{graphicx}
\usepackage{textcomp}
\usepackage{balance}
\usepackage[dvipsnames]{xcolor}
\usepackage{comment}
\usepackage{enumitem}
\usepackage{hyperref}
\usepackage[utf8]{inputenc}
\usepackage{algorithm}
\usepackage{algpseudocode}
\usepackage{amsmath}
\usepackage{titlesec}

\newcommand{\publicationNotice}{
    \begin{center}
        \fbox{
            \parbox{0.9\textwidth}{
                \textbf{Accepted Article:} This preprint article has been accepted by 32nd IEEE International Requirements Engineering 2024 conference, Iceland. 
                \\
                © 2024 IEEE.  Personal use of this material is permitted.  Permission from IEEE must be obtained for all other uses, in any current or future media, including reprinting/republishing this material for advertising or promotional purposes, creating new collective works, for resale or redistribution to servers or lists, or reuse of any copyrighted component of this work in other works.
            }
        }
    \end{center}
}

\titlespacing{\subsection}{0pt}{\parskip}{-\parskip}

\newboolean{showcomments}
\setboolean{showcomments}{true} 
\ifthenelse{\boolean{showcomments}}
  {
		\newcommand{\nbb}[2]{
		\fcolorbox{black}{yellow}{\bfseries\sffamily\scriptsize#1}
		{\sf$\blacktriangleright$\textcolor{blue}{\textit{#2}}$\blacktriangleleft$}
		}
		
		\newcommand{\remarks}[1]{\color{red}[#1]\color{black}}

		\newcommand{\del}[1]{\textcolor{red}{\sout{#1}}} 
  }
  {
		\newcommand{\nbb}[2]{}
		\newcommand{\remarks}[1]{}

		\newcommand{\del}[1]{} 
  }

\makeatletter
\newcommand{\linebreakand}{%
  \end{@IEEEauthorhalign}
  \hfill\mbox{}\par
  \mbox{}\hfill\begin{@IEEEauthorhalign}
}
\makeatother

\newcommand{\lesson}[1]{\vspace{0.5em}\setlength{\fboxsep}{0.015\linewidth}\noindent\fcolorbox{BlueViolet}{white}{\parbox{0.96\linewidth}{\textcolor{BlueViolet}{\textbf{Lesson learnt:}} #1}} \vspace{0.5em}}

\def\BibTeX{{\rm B\kern-.05em{\sc i\kern-.025em b}\kern-.08em
    T\kern-.1667em\lower.7ex\hbox{E}\kern-.125emX}}
\begin{document}
\publicationNotice

\title{Engineering Safety Requirements for Autonomous Driving with Large Language Models
\thanks{Funded by Sweden's Innovation Agency, Diarienummer: 2021-02585}}

\author{
\IEEEauthorblockN{Ali Nouri}
\IEEEauthorblockA{\textit{Volvo Cars \&}\\ \textit{Chalmers University of Technology} \\ \textit{Gothenburg, Sweden} \\ ali.nouri@volvocars.com}

\and

\IEEEauthorblockN{Beatriz Cabrero-Daniel}
\IEEEauthorblockA{\textit{University of Gothenburg} \\ \textit{Department of Computer Science and Engineering}\\\textit{Gothenburg, Sweden} \\ beatriz.cabrero-daniel@gu.se}

\and

\IEEEauthorblockN{Fredrik Törner}
\IEEEauthorblockA{\textit{Volvo Cars} \\
\textit{Gothenburg, Sweden} \\
fredrik.torner@volvocars.com}

\linebreakand

\IEEEauthorblockN{Håkan Sivencrona}
\IEEEauthorblockA{\textit{Zenseact} \\
Gothenburg, Sweden \\
hakan.sivencrona@zenseact.com}

\and

\IEEEauthorblockN{Christian Berger}
\IEEEauthorblockA{\textit{University of Gothenburg} \\ \textit{Department of Computer Science and Engineering}\\ \textit{Gothenburg, Sweden}  \\ christian.berger@gu.se}
}

\maketitle

\begin{abstract}
Changes and updates in the requirement artifacts, which can be frequent in the automotive domain, are a challenge for SafetyOps. Large Language Models (LLMs), with their impressive natural language understanding and generating capabilities, can play a key role in automatically refining and decomposing requirements after each update.
In this study, we propose a prototype of a pipeline of prompts and LLMs that receives an item definition and outputs solutions in the form of safety requirements. This pipeline also performs a review of the requirement dataset and identifies redundant or contradictory requirements.
We first identified the necessary characteristics for performing HARA and then defined tests to assess an LLM's capability in meeting these criteria.
We used design science with multiple iterations and let experts from different companies evaluate each cycle quantitatively and qualitatively. Finally, the prototype was implemented at a case company and the responsible team evaluated its efficiency.

\end{abstract}


\section{Introduction}
Software for Autonomous Driving (AD) is complex and ensuring its safety is critical. It must be assessed throughout the many sub-systems and sub-components that make up the desired AD behaviour, rendering it a difficult and complex task itself. Moreover, the complexity of the environment in which AD systems operate, and the possible malfunctions when interacting with other traffic agents lead to an almost infinite exploration space for potential issues.

Techniques to engineer and maintain requirements for such complex systems are commonplace in industrial setups. An example is Hazard Analysis and Risk Assessment (HARA), based on standards like ISO 26262~\cite{ISO26262} and ISO 21448~\cite{sotif}, is an example to mitigate such issues to identify possible hazardous events and to assess their risk in a systematic way. Various strategies are used to specify safety requirements for events associated with a high risk that will be verified and validated at different stages of the project.

However, function descriptions, operational environments, and regulations in the automotive domain rapidly change. Hence, a company-specific DevOps cycle \cite{devopsgoogle} including HARA needs to be repeated iteratively each time, when a new hazard or scenario is identified to potentially specify new relevant safety requirements~\cite{SEAAAliIndustrial}.
An important ingredient for conducting HARA is brainstorming about possible hazards, which requires imagination and creativity. Recent technological successes in AI such as LLMs might be able to assist engineers when brainstorming.


Our research goal is to design an LLM-based prototype capable of effectively supporting human engineers to specify safety requirements as needed for HARA in the context of complex automotive functions like AD. The design of the prototype was done in cycles: Firstly, identifying the LLM's limitations, followed by focusing on the task breakdown and prompt engineering, and finally evaluating the results in a real-world industrial context. We aim at answering the following research questions:

\begin{enumerate} [leftmargin=0.8cm]
    \item[RQ1] What are the limitations of using LLMs for specifying safety requirements for AD functions?
    \item[RQ2] What is the task breakdown to enhance the LLMs' performance in specifying safety requirements using HARA?
    \item[RQ3] How can prompt engineering enhance the LLMs' performance in specifying safety requirements for AD functions?
\end{enumerate}

Our observations indicate that LLMs have the potential to effectively and efficiently specify safety requirements for AD functions.
The remainder of the paper is structured as follows: Sec.~\ref{sec:method} presents the methodology used to iteratively improve the LLM-based prototype. Sec.~\ref{sec:cycle1} to~\ref{sec:cycle3} discuss and justify the main design changes based on the evaluations of the generated artifacts. Sec.~\ref{sec:discussion} provides an overview of the main design practices needed to support an automation of HARA in real-world settings and discusses associated benefits and risks. Sec.~\ref{sec:conclusion} concludes with final remarks.


\subsection{Background}

An \emph{Item Definition} is a short document on an item that includes a description of its functionality and its interaction with the environment and driver.
HARA uses an item definition to deliver safety requirements called safety goals. Hazard identification is the initial step in HARA achieved by legacy catalogues or systematically through the use of guide-words such as ``omission'' and ``commission''. Omission refers to ``the function does not produce the intended effect'', while commission describes ``The function produces the intended effect when it should not''.

The hazards are combined with relevant scenarios to form hazardous events. 
Systematic identification of such scenarios is crucial for aiming at the completeness argument of identified hazardous events. Scenario catalogues are collected from field data or systematically constructed using scenario factors as mentioned in ISO 21448~\cite{sotif} or suggested by the PEGASUS research project~\cite{Pegasuslayers}. It is crucial to formulate scenarios and their corresponding hazardous events with the correct level of granularity and abstraction~\cite{RolfHARA}. Otherwise, it can lead to an incomplete, wrong, or overly conservative HARA.

Safety goals are then specified if the identified hazardous events assessed as safety-related. The safety goals shall not contain technological solutions, as the technical aspects will be specified in the next abstraction levels such as system level~\cite{SEAASTPA}. 
Common mistakes, such as ambiguous formulations, internal inconsistencies with other safety goals, or duplication of information in safety goals should be avoided.
Ensuring the correctness and completeness of safety goals is crucial as they form the foundation for all other activities prescribed in ISO 26262~\cite{ISO26262} and ISO 21448~\cite{sotif}.

\section{Related Work} \label{sec:RelatedWork}
Prompt engineering techniques such as few-shot learning~\cite{Brown2020} are crucial aspects to obtain useful outputs from LLMs. Studies such as~\cite{promptPatterns2023} are focusing on extracting effective patterns, while other researches such as~\cite{zheng2023chatgpt} have concentrated on limitations of LLMs.
Fang et al.~propose the use of LLMs in anomaly detection in autonomous vehicles from a functional safety, SOTIF, and cyber security perspective~\cite{Fang2024}. The use of an LLM is proposed to predict the error from logged data, diagnostic through rounds of dialog with the model, and being employed as the Human Machine Interface to provide information about anomaly to the user. However, trustworthiness of LLM is raised as a limitation of such technology which needs to be considered. 
LLMs can be used for understanding the reason of actions in a system, although it is not suggested to use them in  autonomous systems or for emergency obstacle avoidance maneuvers~\cite{Lowe2023}.

According to~\cite{Dima2021}, the performance of NLP might drop for technical applications as the volume of domain specific technical data is not comparable with the data that was used to train NLP systems such as GPT. For instance, the grammatical and terminological aspects are different in standard English texts versus industrial, domain-specific texts. The terms used in a requirement can be even be more specialized for one department or even a project~\cite{Bertram2022}. 
This is observed also in our experiments as, for example, ``the system should avoid the malfunction\dots'' might be acceptable from grammatical point of view but not according to a company language. 
This is why~\cite{Dima2021} propose a human-in-the-loop iterative tailoring of NLP, so called Technical Language Processing (TLP). The industrial standards and technical dictionaries can be used as data sources. 

Bertram et al.~proposed using few shot learning to develop a domain specific language (DSL) for the automotive domain to translate legacy requirements from other departments or projects for a new project and also to be used as a correction system when formulating new requirements~\cite{Bertram2022}. They used GPT-J6B, which according to the authors was the best model publicly available at that time in combination with a set of publicly available data set of requirements. 

Some studies have focused on using LLMs as assistance for engineers to perform safety analysis methods such as System Theoretic Process Analysis (STPA)~\cite{AEBGPT} or hazard analysis~\cite{diemert2023large}. In these studies, the engineer continuously interacts with the LLM to obtain the desired output but automation was not a primary goal.

\section{Methodology: Design Science} \label{sec:method}

\begin{figure*}
    \centering
    \includegraphics[width=.9\linewidth]{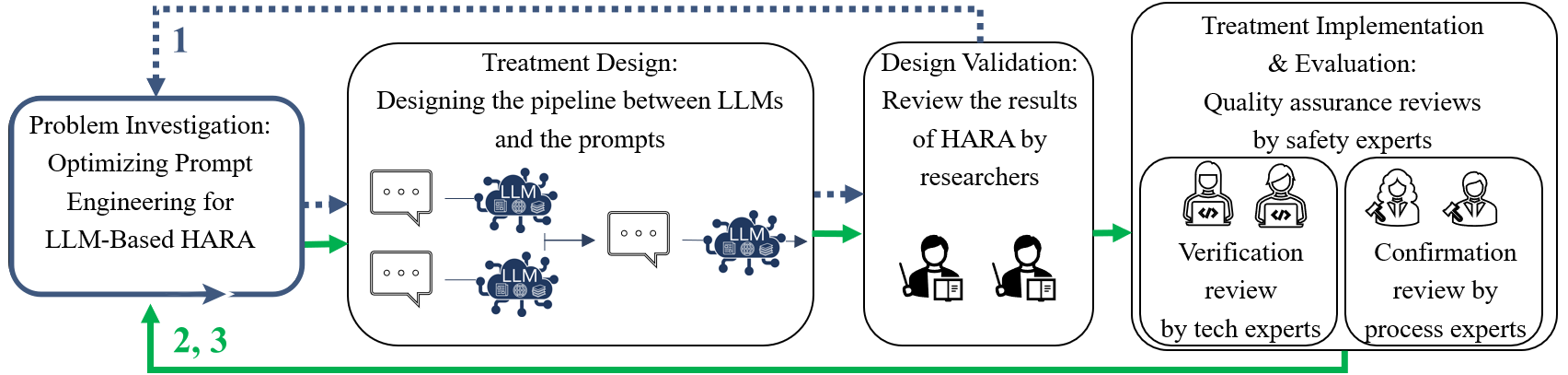}
    \caption{The design and engineering cycles used for our study: The dotted blue arrow is showing the design cycles, which the first one is reported (No.~1). The green arrow is showing the engineering cycles, which is done in two cycles and the results are reported (No.~2\&3).}
    \label{fig:DesignEngineeringcycle}
\end{figure*}

We adopted the Design Science Methodology (DSM) as outlined by Hevner et al.~\cite{DesignScienceHevner} and  Wieringa~\cite{DesignScienceRoel} to design and evaluate an LLM-based prototype capable of automatically specifying safety requirements for AD functions.

\subsection{Problem Identification} \label{sec:problem} 

We used the following query prior to the first design cycle to retrieve peer-reviewed papers from Scopus, IEEE, and ACM about evaluation studies for the use of LLMs for safety requirements in the automotive domain:

\begin{verbatim}
((Safety OR Safe) AND
Requirement AND Automotive AND
((Large Language Model) OR GPT OR LLM))
\end{verbatim}

Relevant related publications, including guidelines and standards to understand industrial practices related to specifying requirements with a special focus on safety, were additionally explored. A summary of the gathered insights about the use of LLMs within Requirements Engineering (RE) and safety assessment is provided in Sec.~\ref{sec:RelatedWork}. Relevant driving scenarios and generic malfunctions were extracted from the literature to serve as a basis to design the initial prototype version, as detailed in Sec.~\ref{sec:cycle1}. 


\subsection{Summary of Research Cycles} 


The LLM-based prototype was designed in three cycles, each consisting on a design phase, an implementation phase, and an evaluation phase. Internal evaluations were conducted additionally as shown in Fig.~\ref{fig:DesignEngineeringcycle} to iteratively improve the prototype within each design cycle~\cite{DesignScienceRoel}. In the last two iterations that we refer to as the engineering cycles, we included independent industrial experts to get feedback on the prototype for improving the design\footnote{Providing study protocol, \url{https://zenodo.org/records/10649245}}.


The findings from each design cycle contribute to the design of the prototype in a different way:
In the \textit{first design cycle}, tests were devised to identify the limitations of LLMs in conducting HARA to specify safety requirements to address \textbf{RQ1}. A first version of the prototype was created based on the insights in white and grey literature, including generic industrial standards for HARA~\cite{ISO26262}. The prototype's generations were internally evaluated as presented in Sec.~\ref{sec:cycle1} in terms of their readability and relevance to automotive and safety.

Based on the identified LLM limitations, the \textit{first engineering cycle} focused on addressing these limitations by breaking down the HARA into LLM-manageable tasks to address \textbf{RQ2}. The tasks constitute a pipeline capable of automating the process as outlined in Sec.~\ref{sec:cycle2}. We reached out to independent safety experts from the automotive industry to evaluate the LLM-generated HARA.

We refined the prototype in the \textit{second engineering cycle} as detailed in Sec.~\ref{sec:cycle3} to improve the quality of the generations based on the received expert feedback. On the one hand, prompt engineering was used to improve each of the tasks to address \textbf{RQ3}. On the other hand, the pipeline was further refined to improve our insights obtained for\textbf{RQ2}. In this last cycle, we studied the integration of the prototype in a real-world industrial context. We used a case company's in-house LLM and reached out to their safety experts to evaluate the LLM-generated HARA for a company's automotive function in its industrial context.



\underline{\emph{Ethical Consideration:}} The ethical principles for software engineering interview studies, including consent, anonymization, and confidentiality, are followed in this study as recommended by Strandberg \cite{swethic}. Each step of the study was reviewed against the ethical checklist.

\section{First Design Cycle: Identify LLM Limitations} \label{sec:cycle1}

\subsection{Artifact Design: What to Automate?} 

Previous work has looked into generating STPA using ChatGPT~\cite{AEBGPT} through a number of user interactions with the system until satisfactory results are provided by the LLM. In contrast, the aim of the present study is to develop an LLM-based automatic tool for HARA. 
In this approach, users provide only the item definition of any safety-critical automotive function to generate a set of safety goals for avoiding or mitigating potential hazardous events of the function. This is then presented in a format easily readable by humans, i.e., a table. Then, the safety goals and HARA will be reviewed by the experts, and necessary improvements will be identified for implementation by the engineers.


\subsection{Artifact Implementation Using ChatGPT} 

The first cycle consisted on a feasibility study to answer the question, ``Can LLM perform safety analysis, and if so, is it useful at all?'' The prompts were iteratively improved by the research team, who had previous experience in HARA, by testing a number of prompt engineering techniques. 

Each prompt contains the relevant process of that specific subtask, which was provided to the LLM, and then the output was used in the next subtask without any modification. The process in the first engineering cycle, including this design cycle, is derived from relevant international standards.
No company-specific examples or detailed instructions were provided to the LLM to avoid biasing the tool. 
Instead, the LLM used the references on which the industry has a consensus, thereby enabling experts from different companies to review and assess it effectively. Moreover, it enables other companies to use the same prompt, which contributes to the generalizability of the study.

Often, company catalogues and international standards provide scenario and malfunction descriptions. Although, we envisioned the tool to be independent and autonomously generate relevant scenarios and malfunctions.
In this manner, we could assess the capability of LLMs in generating relevant scenarios and malfunctions by using only the function description and a generic process, as discussed in Nouri et al.~\cite{AliLLMHARA}. 

Inputs often rely on figures with dense information, difficult to transmit in words. Some tests were conducted to evaluate whether the LLM could interpret the scenarios provided in this manner, such as figures.

\lesson{However, the results showed that the technology is not yet mature enough to interpret correctly other ways of transferring technical information, such as figures. For these reasons, plain text was selected to represent scenario information, and the diagrams in item definition.}

To test the designed solution, a function was selected to be analysed. Two options were considered: (i) a well-known function with publicly available documentation, such as Autonomous Emergency Braking (AEB); and (ii) a novel function, Collision Avoidance by Evasive Maneuver (CAEM), that, to the best of the authors knowledge, is currently less mature than AEB in terms of publicly available materials.

\begin{figure}
    \centering
    \includegraphics[width=1\linewidth]{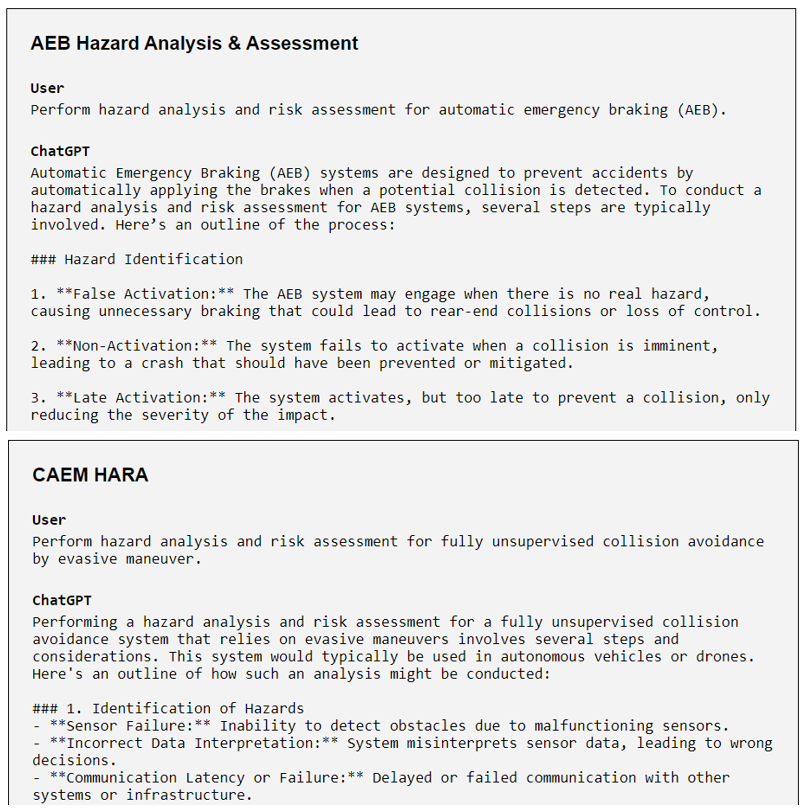}
    \vspace{-.6cm}
    \caption{Figure 2: The first three hazards in two generated HARAs, showing the LLM's prior knowledge about AEB in contrast to CAEM.}
    \label{fig:AEBvsCAEM}
\end{figure}

In the initial tests, the LLM's prior knowledge of HARA and automotive functions was assessed by asking it to analyse two functions without providing a function description or instructions for HARA, as shown in Fig.~\ref{fig:AEBvsCAEM}. On the one hand, a well-known function with publicly available documentation, Autonomous Emergency Braking (AEB) was tested. A second function, Collision Avoidance by Evasive Maneuver (CAEM), that, to the best of the authors knowledge, is currently less mature than AEB in terms of publicly available materials, was also used to test the LLM's prior knowledge.

The generated HARAs for AEB and CAEN, partially shown in Fig.~\ref{fig:AEBvsCAEM}, showed differences in the LLMs knowledge. For AEB, the hazards can be mapped to ``Commission'', ``Omission'', and ``Delay'' and are in the correct abstraction level. However, that was not the case for CAEM, leading to an incorrect results. 

\lesson{Given that the tool should generalise to any safety-critical function, this experiment confirmed the importance of selecting a novel function for evaluating the LLM capabilities for performing HARA. This raises concerns about the validity of experiments that compare the results generated by LLMs against well-established, publicly available baselines.}

This might, however, be the case for the chosen GPT model and its current version. Other models, which are trained on different data, might exhibit different performance. 

\subsection{Artifact Evaluation: Internal Assessment} 

The LLM-generated HARA for CAEM was internally evaluated, and the issues linked to the prompts, which were iteratively improved to guide complete HARA generation. The needed detail in the description of each each HARA sub-task soon increased to the point where decomposition of the steps was necessary, as discussed in further detail in Sec.~\ref{sec:cycle2}.

For instance, specific analytical capabilities, such as kinematic estimation in a scenario, are needed for HARA, since estimating the position of each agent and the impact speed during a potential collision would be an essential factor in assessing risks. Specific instructions where then crafted to guide the LLM in this task, and the prompts were tested multiple times with different temperatures for the GPT model.

Other issues were related to the specific meaning of some words in automotive safety engineering.
For instance, the usage of modal verbs: in automotive safety RE ``should'' reports recommendations, while ``shall'' indicates mandatory requirements~\cite{ISO26262,Bertram2022}. 
The LLM was also questioned for the definitions, in the scope of an automotive context, of key terms such as ``scenario'', ``malfunctioning behaviour'', ``hazard'', ``severity'' or ``safety goal''. The results, reported in the Appendix\footnote{Providing sample test on LLM, \url{https://zenodo.org/records/10644052}}, showed that the definitions provided by the LLM were either not correct or too generic. 

\lesson{Therefore, it was critical to provide definitions for key terms as part of the prompting strategy.}

\section{First Engineering Cycle: a Pipeline of Prompts} \label{sec:cycle2} 

\begin{figure*}
    \centering
    \includegraphics[width=1\linewidth]{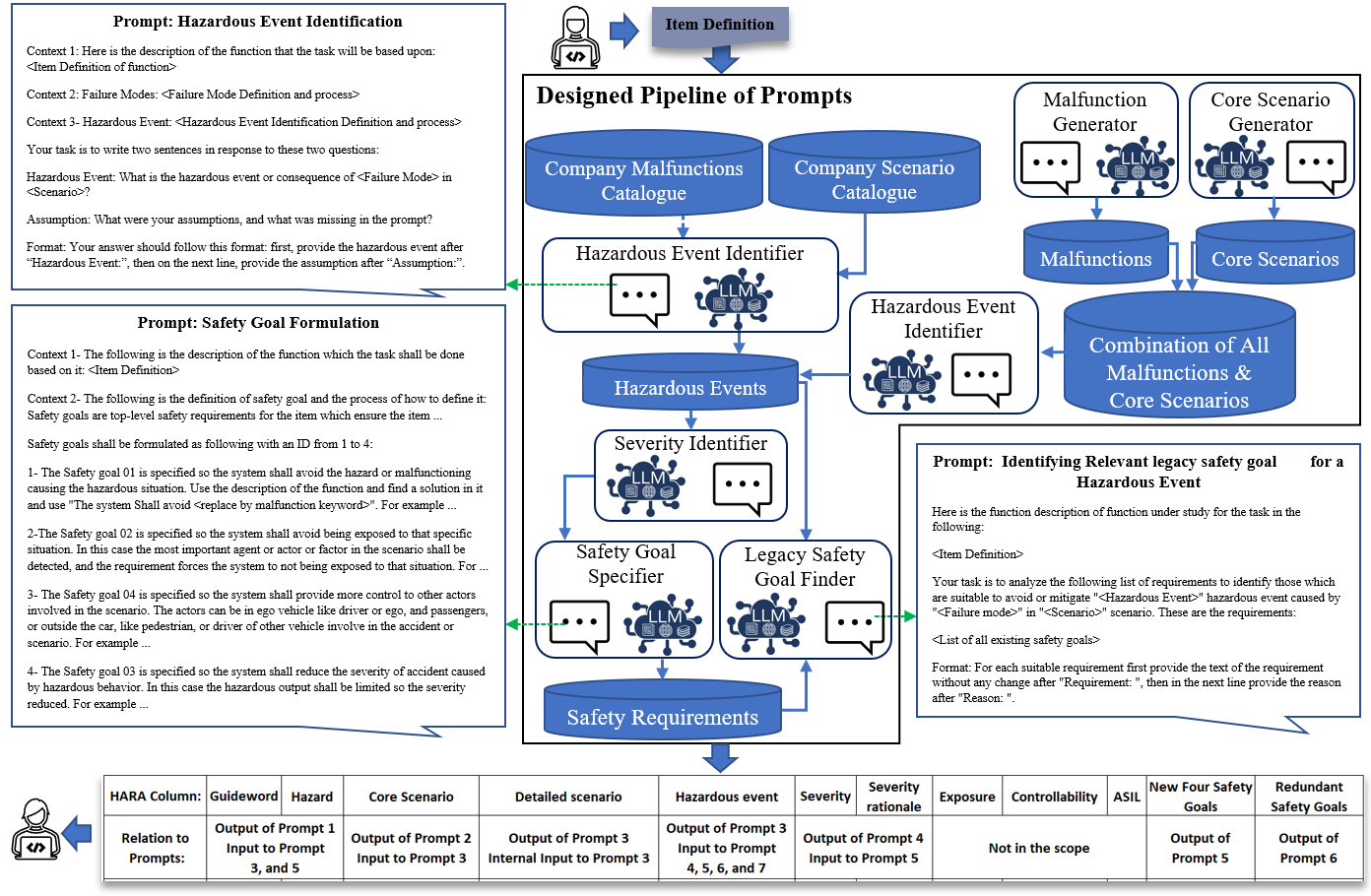}
    \caption{Final version of the pipeline for the LLM-based tool for HARA: The generations are stored and transmitted between sub-tasks automatically using a Python script, without human intervention. In the end, the specified safety requirements are provided in a human-readable table for expert review.}
    \label{fig:HARAPipeline}
\end{figure*}

After the initial feasibility study and design phase as described in Sec.~\ref{sec:cycle1}, an engineering cycle started that involved external experts with diverse backgrounds and from different companies.

\subsection{Artifact Design: Task Breakdown} 

The experiments in the first cycle showed that HARA has to be broken down into tasks that require specific information. Each task is meant to generate or refine a particular artifact that becomes the input for the next sub-task. The pipeline is presented in Fig.~\ref{fig:HARAPipeline} and further details about the task breakdown, including prompt engineering, are reported in~\cite{AliLLMHARA}.

The complete analysis should then be performed automatically without any intervention for allowing human experts to review the final results. As it is shown in Fig.~\ref{fig:HARAPipeline}, the LLM-based HARA is designed to be automated without an engineer's intervention, where the input is the item definition (i.e., the function description), and HARA results are the output. Moreover, the stochastic behaviour of LLM usage shall be avoided in some sub-tasks such as replacement of the varying factors like malfunction or scenario.

\subsection{Artifact Implementation: GPT API Automatic Calls} 

In the second cycle, the OpenAI API for \textit{GPT-4.0} was used to perform each of the steps in the pipeline in a fully automated manner without intervention. For each task in the pipeline, the LLM therefore receives a prompt to generate or refine an artifact that is in turn retrieved with a Python script and stored in a plain text file. 

At the end of the pipeline, the result is a table with the complete HARA, which allows humans to review the complete HARA artifact. Without accounting for the human expert reviews, which necessarily take time, generating the HARA for the selected function can be completed by the LLM-based tool in less than a day, a far shorter time than what is typically required by a team of experts. 

\lesson{Some of the tasks in the pipeline, such as methodically combining scenarios and potential malfunctions, require completeness but not creativity. Rule-based approaches are therefore more convenient, while LLMs can expand them or provide natural language descriptions.}


\subsection{Artifact Evaluation: Interviews with Experts} 

In order to empirically evaluate the proposed artifact, which includes the pipeline and the prompts for each task, we engaged nine automotive safety experts to evaluate the resulting HARA for the selected function (CAEM). The experts were selected from a pool of experts with experience in performing or reviewing HARA for AD and ADAS functions and are knowledgeable about CAEM. They are from three different AD development companies and accumulate on average more than ten years of relevant experience (minimum of five years).

The HARA evaluation typically involves a Verification Review (VR), concerning correctness of technical and project related aspects, and a Confirmation Review (CR), concerning the correctness of process related aspects. In the first step, a package including the item definition, HARA, a review checklist, and a review comment template is provided for review. Experts were not informed whether AI or humans conducted the analysis to avoid biasing them against or in favor.
In the second part, a follow-up meeting is held with each expert during which they clarify their comments. Then, they are asked to provide their overall judgment for each criterion. Finally, they are informed about the LLM-based HARA pipeline and were invited to offer any proposals they might have for improvements.


The review comments were gathered, stored, and analysed. A follow-up meeting was then held with each expert during which they could clarify their comments. Then, they are asked to provide their overall judgment for each criterion. A follow-up meeting is arranged for each expert to clarify their comments. We received 71 review comments from the nine experts who are asked to review the HARA for 20 hazardous events ~\footnote{Appendix provides part of the LLM-based HARA without human intervention,~\url{https://doi.org/10.5281/zenodo.10522786}}. Some were not related to the scope of this study and the rest are clustered in 5 major categories based on the root causes. A representative sample of the reported review comments is presented in Table~\ref{tab:ReviewComments}. Some of these comments are generic, while others are specific to one or several rows in the HARA. 

\begin{table*}[]
    \centering
    \caption{Summarize the clustering of review comments based on root cause. The review focuses on the HARA results of the developed prototype for CAEM in the first engineering cycle. IDs refer to the Hazardous Events provided in the appendix.}
    \label{tab:ReviewComments}
    \begin{tabular}{|p{2cm}|p{15cm}|r}
        \hline
        \textbf{Root cause}  & \textbf{Sample review comment} \\
        \hline
Scenarios& 
CR3: ``The detailed scenarios are too detailed, which results in very specialized scenarios while excluding many other scenarios and also a risk of artificially lowering of E.'' \\
 & CR3: (ID 50) ``Ensure consistency: In detailed scenario there is a ``truck approaching from the left'' and in the severity rationale the truck is ``a large stationary object'', this is inconsistent.'' \\
 &  VR2: (ID 22) ``Better to use VRU instead of Pedestrian in order to cover wider range of unprotected road users'' \\
 &  CR2: (ID 51, 77, 91) ``The scenario is too unclear to be able to formulate a valid hazardous event'' \\
 
\hline
Hazardous event & CR2: (ID 19, 38, 25, 111) ``The Hazardous event does not correlate with the malfunctioning behaviour in the described scenario''  \\
\hline
Scenario & VR5: `` ... difficult to determine completeness. Was a systematic approach applied? ... ''  \\
 Completeness&   VR3: ``Have we covered the sharp turns, when commission has happened. That might lead to lateral instability.'' \\
\hline
Severity& CR3: ``Not enough rationale provided for the stated S. Many assumptions made without proper rationales.''  \\
Identification& VR5: (ID 17) ``CAEM doesn't seem to be limited in speed. How was S2 determined, vehicle speed could have been 130 kph? ...''\\
 &  CR3: (ID 77, and 196) `` In severity rationale it is stated `max allowed speed' in ID 77 and `maximum allowed speed' in ID 196. This is not defined and if there is an upper limit of the host vehicle speed for CAEM this could be a safety mechanism.'' \\
\hline
Safety Goals  & VR8: ``Some safety goals have large overlap, ... Consider generalizing ....''  \\
 Formulation& VR5: ``Safety goals do not need to explain why they exist, like ``...  to prevent unnecessary lane changes''. ... specify the goal, such as ``CAEM shall not cause lane departure unless to avoid collision''. '' \\
 & CR3: ``Ensure unambiguous safety goals: The safety goals contains a lot of undefined parts.'' \\
 & VR6: ``Vicinity need to be precise. The invitation shall be in in case the collision is imminent in-front. ...''\\

 & VR1: (ID 22, 23, and 25) ``Safety Goal 23 is more general and it includes Safety Goal 22. 
Safety Goal 25 is similar to Safety Goal 23'', ``Many safety goals are referring to same thing but different phrasing. '' \\ 
 & VR2: (ID 22) ``when necessary" is vague and ambiguous.'' \\
\hline
    \end{tabular}
\end{table*}

Only a sub-set of the generations could be reviewed due to the large amount of relevant safety requirements for an AD function. A minimal selection was created by clustering scenarios and malfunctions to test the pipeline on. The selection is made by the LLM, through an instruction to select diverse samples to better capture the systematic failures of the prototype.

A checklist extracted from ISO 26262~\cite{ISO26262} was used to evaluate the selected requirements; although interviewees were permitted to include their own criteria and justifications. The provided checklist is as follows:

\begin{enumerate} [leftmargin=0.5cm]
\item [a.] Considering the Operation Design Domain and the output under analysis (i.e., lateral motion request), have all failure modes or functional insufficiencies (i.e., Commission and Omission) been identified in the HARA?
\item [b.] Have all relevant hazardous events been identified? (e.g., the relevant scenario elements in ODD are covered for both Omission and Commission)
\item [c.] Have all Hazardous Events been correctly formulated to present the consequence of the identified malfunction (or functional insufficiency) in the specified scenario? 
\item [d.] Are the safety mechanisms excluded from the analysis? (e.g., no assumption is made on the possible internal mechanisms to avoid the hazardous events)
\item [e.] Are all assigned severities corresponding with the rational?
\item [f.] Is there any inconsistency within the results of HARA? (e.g., the severity classifications are different for the same consequence.)
\item [g.] Are there safety goals formulated for each hazardous events with severity higher than S0?
\item [h.] Does the safety goal cover the hazardous event? (i.e., the safety goal is enough to avoid or mitigate the hazardous event) 
\item [i.] Have all safety goals been formulated in a correct way? (e.g., Unambiguous)
\item [j.] Is the HARA contributing to the achievement of functional safety or SOTIF?
\end{enumerate}

\begin{figure}
    \centering
    \includegraphics[width=1\linewidth]{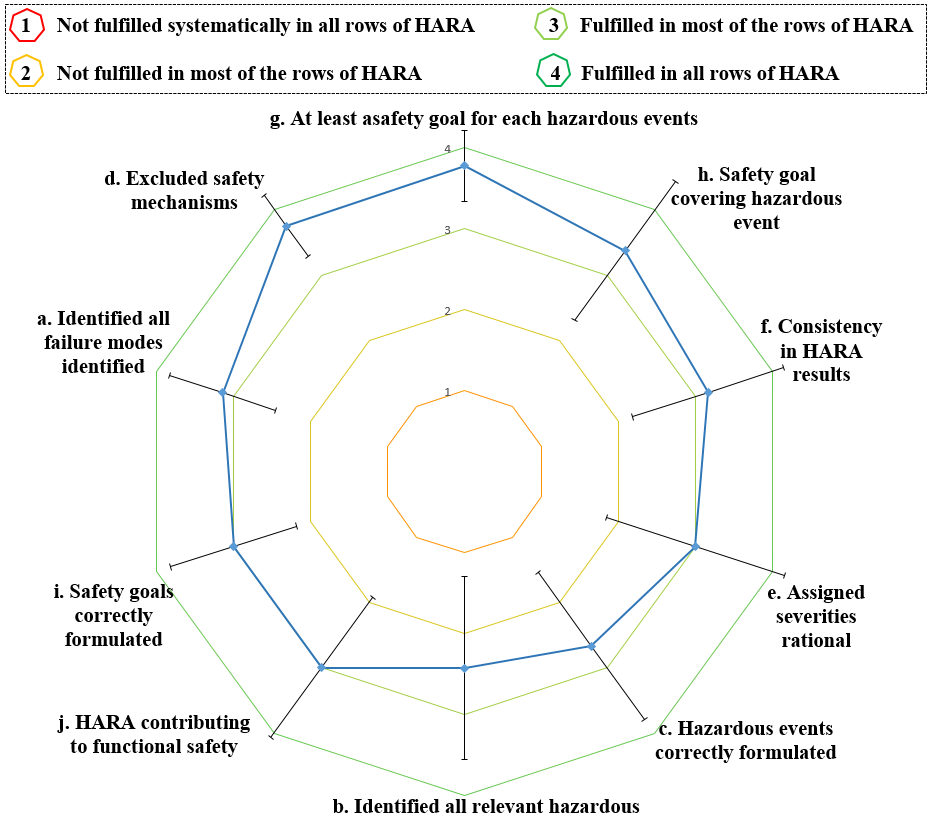}
    \caption{Average expert scores for HARA criteria. Criteria are ordered to display the highest scores at the top and the lowest at the bottom, highlighting the relative scores.}
    \label{fig:spider}
\end{figure}

Then, they are asked to provide an overall score conclusion for each criterion in the checklist. The experts are provided with five options to answer ``To what extent the provided HARA satisfy each criteria?'' and then each criterion is provided to them. The '3- No Opinion' option was first excluded, followed by updating the range from 1 to 4.  Then the average and standard deviation are calculated. Fig.~\ref{fig:spider} presents a radar chart of the average scores given by experts for each criterion, along with the calculated standard deviation. The criteria were reordered to display the highest scores at the top and the lowest scores at the bottom. 

\begin{enumerate}
\item [1:] Not fulfilled systematically in all rows of HARA
\item [2:] Not fulfilled in most of the rows of HARA
\item [3:] No opinion
\item [4:] Fulfilled in most of the rows of HARA
\item [5:] Fulfilled in all rows of HARA
\end{enumerate}

As shown in Fig.~\ref{fig:spider}, experts agreed on the fulfillment of criteria, addressed in question ``g'' and ``d''. Please note that, related to criterion ``g'', it is the rule-based part of the pipeline (i.e., a Python script) that is responsible which is calling the API for each hazardous event with a severity higher than S0. Moreover, criterion ``i'', concerned with the quality of the safety goals, was also fulfilled, based on the average score represented in Fig.~\ref{fig:spider}.


\subsubsection*{Areas for Improvement} 

Although no average score fell below ``2- Not fulfilled in most of the rows of HARA'', there was disagreement between the experts regarding questions ``b'' and ``c'', related to identifying and correctly formulating relevant hazardous events. Such differences in scores underscore the subjective nature of expert judgement, which is often influenced by individual experiences.
For both ``b'' and ``c'', two experts reported that the AI was not successful in all rows in the HARA.

For criterion ``b'', one expert, who voted for ``1- Not fulfilled ...'' commented, ``Difficult to understand the approach (used to derive the scenarios) to investigate scenarios, and \emph{difficult to determine completeness}. Was a systematic approach applied? ...''.
After being informed about the LLM-based approach and being asked for possible remedies, the \textbf{same expert} mentioned, ``... \emph{it (i.e., scenario identification) is more like a craft and no systematic way (exists) till now.} ...''.

Our interpretation is that although there is a need for a systematic approach to argue for completeness, since no such method currently exists, the final completeness argumentation should be formed \textbf{case by case for each function and HARA}.

Moreover, one possible root cause for missing some scenarios might be the clustering and filtering in the last step. It can potentially be improved by merging and abstracting of the scenarios in each cluster, instead of selecting some per cluster.

For ``c'',  the comment `` Some safety goals have large overlap, ... Consider generalizing .... '' is one example. This issue arises from formulating the safety goals in separate conversations, leading to different formulations of the same requirements.
A possible remedy is to cluster the requirements, merge, and then abstract from them, or use a buffer to record previously formulated safety goals and recommend the model to first check these before formulating new ones if necessary.
Some reviewers commented on formulation of safety goals such as ``Ensure unambiguous safety goals ...''.

Using these comments as lessons learned in the prompts, by identifying forbidden patterns or providing wrong/correct examples, might increase the quality of the safety goals, as examples typically improve quality.


Finally, they are informed about the LLM-based HARA pipeline and invited to offer any proposals they might have for improving it.
Two experts who gave the lowest average score gave the following comments after they are told it is an LLM-based HARA:
\begin{itemize}
    \item ``We can't rely on it but it is ok to use it as a guide. As a tool maybe helpful and useful as a help.''
    \item ``[\dots] can be a complement but hard to replace current way of doing''
\end{itemize}

\lesson{These results suggest that while LLMs and current prompts are not sufficient on their own yet, they are valuable as a preliminary step in a HARA, facilitating faster development of valid versions by development engineers.}

\section{Second Engineering Cycle: Refining Pipeline and Prompts, Using an In-house LLM} \label{sec:cycle3}

We analysed the review comments and identified potential root causes for each category of comments from the first engineering cycle. Then, different techniques were used to improve the results, although some limitations are due to the inherent limitations of the technology such as risk for fabricating false information or hallucinations.

\subsection{Artifact Implementation: Using an In-house GPT Model} 


The API used in the second engineering cycle is the case company's GPT-4 running as Azure OpenAI Service.
Since the model is protected with respect to the usage of intellectual property, we detailed each prompt using the internal process description of the case company. Additionally, we used inputs from a novel automotive function from the case company. As the novel automotive function's item definition is not publicly available and hence, could not be scraped during the training phase of the LLM, and a baseline of the function exists within the case company, we were able to compare the results of the prototype with the internal baseline.

\subsection{Artifact Design: Prompt Engineering}

\subsubsection{Explainability} The team that performed the HARA and the reviewers discuss the comments in multiple workshops. Since this is not possible with LLMs, there is a need to ask an LLM to attach additional explanations to each output such as background information or reasoning. Recent research in this area is suggesting the use of retrieval-augmented generation (RAG) to ground LLM-generated output in a facts database. 

\lesson{Background, reasons, or assumptions attached by the LLM to the final output help a human reviewer to better understand the output. They can also serve as context for the next LLM in the pipeline.}

\subsubsection{``The scenarios are extremely detailed''}
Some comments referred to the level of granularity in the scenario description. 
We did not perceive this as a weakness; the prompts are designed to detail the scenario, thereby leaving fewer assumptions for the next LLM in the pipeline.
For instance, the system uses the scenario to identify the severity of the hazardous event; missing details leave room for assumptions by the LLM, which reduces the accuracy and explainability of the decision.
However, we also concur with the reviewers that this is not the typical approach to conducting HARA.
One potential solution is to filter the detailed scenarios and cluster them under each core scenario.

\lesson{Some restrictions in the process were established to better match human weaknesses, such as limited time, and strengths, such as access to context. If we decide to replace some activities with LLMs, there might be a need to adapt some of those processes and restrictions to better match the strengths and weaknesses of the system.}

However, any changes to the process should be made and analysed carefully to not compromise the safety of the process's final outcome.

\subsubsection{Detailing the process}
A cluster of comments was related to some inaccurate outputs at each step, such as ``Hazardous event does not correlate with the malfunctioning behaviour in the
described scenario''.
As the process defined in the standards lacks details, explaining each step in more detail is seen as a potential solution to improve the results.
This was achieved by using the company's well-detailed internal process, which significantly improved the outputs.

\subsubsection{Examples}
Using few-shot learning~\cite{Brown2020} is seen as an effective way to improve the generated output; therefore, generic examples are used to explain the process, which are not specific to the function under analysis. These examples are hard-coded into the prompt since they are not function-relevant.


\subsection{Artifact Design: Safety Goal Engineering}

To specify the safety goals in the initial engineering cycle, the prompt contained only the process of how to specify a safety requirement that could pass quality gates. However, defining the technical aspects of a safety goal was left to the model's discretion. In the second engineering cycle, we outlined four different strategies for the system to guide the LLM on the technical aspect as well.

Each hazardous event can be avoided in multiple ways; here, we specify four main strategies to avoid it. These strategies would lead to four safety solutions (i.e., safety goals) per hazardous event. Then the engineers need to decide which safety solution is a better match capability of technology, and their current system. The engineers might select multiple solutions for each hazardous event to improve the safety of the system.

\textbf{``Avoiding failure mode''} would be the most natural requirement, as it attempts to eliminate the main cause of a hazardous event. If technological limitations permit, this is the most desirable safety goal since it does not limit the function. However, this is not always possible, especially with immature technologies and functions, which necessitates having other safety goals, as explained in the next strategies.

\textbf{``Avoid being exposed to the situation''} is one possibility, known today as the Operational Design Domain or ODD. This strategy assumes that confidence in the first strategy is low, so it is conservative to avoid the scenario as the second element in the hazardous event. For example, if confidence in detecting pedestrians is low, then scenarios with a high exposure to pedestrians would be removed from the ODD.

\textbf{``Improve controlability''} for any road users involved in the hazardous event can be seen as a parallel safety requirement. This is not limited to the driver or passengers of the vehicle but also extends to other road users such as pedestrians or other drivers.

\textbf{``Reduce the severity''} is a strategy where the LLM is tasked with identifying relevant factors in the hazardous event that can decrease severity. These may include limitations on speed, acceleration, deceleration, or lateral motion. This strategy can be employed in conjunction with other strategies to mitigate risk.

The requirements formulated by each strategy are intended not only to be used directly in the design, if approved, but also to serve as a helpful brainstorming tool for designers to think outside the box.



\subsection{Artifact Design: Pipeline Improvements} 

\textit{Safety Requirements Redundancy Finder:}
According to ISO 26262, safety requirements must not contain duplicated information, which means that the information in one safety goal should not be repeated in other safety goals. As noted by the experts in the first engineering cycle, there are multiple requirements containing the same information but with different formulations.

The main cause of this issue is that the safety goals are formulated in separate prompts, and as a consequence, the model is not aware of the existence of already existing requirements. To address this, we implemented an extra step that checks the already existing safety goals and identifies all relevant safety goals that have been specified to avoid redundancy. Then, if the engineer is not satisfied with the existing safety requirements, they can select one that the LLM has specified as a new requirement.



\subsection{Artifact Evaluation: Case Study} \label{sec:Case_Study} 

We presented our prototype to the team responsible for performing a HARA for a novel function. Then, four team members are asked to participate in separate individual sessions to demo the tool. The users are asked to provide failure modes and scenarios they wish to test the prototype on, and then they are asked to review the results. The sessions lasted between 1 to 2 hours. Subsequently, they are asked the following questions, and their answers are recorded and anonymized.

The experts were then asked to discuss the validity and formulation of the safety goals provided by the tool, in terms of meeting the necessary safety standards or regulations. Moreover, they were asked to compare in their formulation to their understanding of ``properly structured safety goals.''
In general, the experts reported that safety goals were structured properly and considered them to be adequate for the \textit{first version} of the tool. Moreover, they considered that the prototype can also facilitate the writing of safety goals.

When asked about the instructions for the formulation of safety goals, the experts stated that they seem useful although, in some instances, the LLM does not follow the instructions, resulting in irrelevant safety goals.
The search strategy for existing safety goals is also seen as a proper step. One participant even suggested expanding the use of this concept to identify redundant requirements between safety and cybersecurity.

On the one hand, the experts said that some safety goals fully address the hazardous events. On the other hand, there were cases that did not completely cover the hazardous events, and the safety goal needed some improvements.

\subsubsection*{Areas for Improvement}

Some safety goals are not seen as relevant, both among the newly formulated safety goals and within the found legacy requirements. One potential solution is to ask the LLM to filter the answers. However, this increases the risk of filtering out relevant ones, since filtering requires engineering judgment, an area in which the LLM is not sufficiently reliable. Therefore, there is a need for a balance between false positives and false negatives. As one of the participants mentioned:
``Giving more options can allow the human to filter and choose the relevant ones. This is better than giving less answers with the risk of missing some. ...''. So our conclusion was to leave the filtering to humans, although it is time-consuming for engineers to read and filter.

The reformulation of the scenario and failure modes provided by engineers is proposed as a potential solution to improve the generated output. However, it needs to be done carefully since, in most cases, the inputs come from company catalogues, which should not be changed.

As we observed, all participants utilized the additional explanations in the background, context, or assumption sections of the answer to better understand the output. One of the participants highlighted the need for more information in the assumptions as a potential improvement.
Multiple participants proposed using the existing data and legacy requirements within the company as a potential solution to improve the prototype's performance.

\section{Discussion} \label{sec:discussion}

This study focused on designing a prototype for using LLMs in HARA. The goal the system was to support human engineers in specifying safety requirements for autonomous driving functions, which is a challenging and time-consuming task.

In the first design cycle, the limitations of using an unconfigured LLM for safety- and automotive-related RE were studied to address \textbf{RQ1}. The findings show that while LLMs are capable of providing definitions and reason about safety-critical software systems, they often lack the specific knowledge that is required to conduct a HARA that needs to be provided as part of the prompts (\textbf{RQ3}). Moreover, the data used to train LLMs, which is usually unknown to users, clearly affect their ability to discuss technical topics related to AD. 

Undesired hallucinations in the generations (\textbf{RQ1}) appeared as another significant limitation of using LLMs for safety RE throughout our experiments. This is a challenging trade-off because the HARA requires allowing for some degree of creativity, which can only be partially controlled by the temperature parameter, but which might also lead to irrelevant or undesired results. Nevertheless, this is not critical as the evaluation of HARA, both VR and CR, is an integral part of the process under all circumstances.

These considerations were used to break down the HARA into tasks, for which specific context and guides could be provided to guide the LLM as presented in Sec.~\ref{sec:cycle2} (\textbf{RQ2}). Some of these tasks as well as the connection between them that we outlined in the pipeline in Fig.~\ref{fig:HARAPipeline} were automated using rule-based systems scripted in Python instead of relying on an LLM. This is done in order to prevent undesired hallucinations in tasks, where neither creativity nor natural language understanding are needed.

The pipeline and prompts were iteratively improved in the subsequent design and engineering cycles to address \textbf{RQ2} and \textbf{RQ3}. As highlighted in the literature, prompt engineering greatly affected the performance of the prototype both in terms of content (i.e., hazardous events, risk assessment, specified safety requirements, etc.) and form (i.e., output format, readability, etc.). 
In the last cycle, we evaluated the prototype after a number of improvements based on the feedback of independent safety experts in a real-world context. For that, the case company's in-house LLM was used to generate the HARA of one of their functions to be compared with the baseline of the function existed within the case company. This helped the reviewers to compare the LLM-generated results to the existing HARA.

The prototype showed promising capabilities throughout the cycles of our study. The interviewed experts stated that the tool was useful and LLM-generated safety requirements were comparable to human-written results. One interviewed expert who preferred human written safety requirements mentioned: ``in the future, it will be a powerful tool that might exceed human competence.'' 
According to the expert, this technology could also increase efficiency of DevOps cycles because ``with some improvements and tool's training, it will optimize the HARA creation time and be a great help in the future.'' 

While the findings reported in Sec.~\ref{sec:cycle2} and~\ref{sec:cycle3} demonstrate the potential of LLMs for the specifying and specifying safety requirements for AD functions in a real world, industrial context, it is important to keep in mind the difficulty of leashing these tools to fit the specific needs of a real-world context. 
This study, though, discusses some design practices that can be used to iteratively improve both the pipeline and the individual prompts, and reports safety experts' views to support the decision.  


\section{Threats to Validity}
We address relevant threats to validity to our study in the following.
\label{sec:Threats}

\emph{Construction Validity:} 
The evaluations in each engineering cycle were conducted by different independent experts, which might have introduced differences between cycles. However, the reviewers in the first engineering cycle could not assess the HARA in the context of the case company. Another reason to use a different set of experts is avoid their bias towards LLM-generated safety requirements after being told in the first engineering cycle. In order to further spot any significant differences that would affect the validity of the study, the interview protocol for each cycle was evaluated through a pilot study.

\emph{Internal Validity:} 
The selected experts, safety engineers with an average of more than ten years of experience, were from three companies involved in the development of AD and ADAS. They had experience with HARA, having performed or reviewed it, which is making them familiar with the review process. 
Although the experts were transparent in their reviews, we implemented two steps to prevent biasing or affecting their opinions, thereby ensuring the reliability of the results. Firstly, we did not specify who performed the HARA and how it was done. Secondly, we provided the checklist as it would be for a HARA performed only by human beings, ensuring it was given to the experts in advance with time for review.

\emph{External Validity:} 
This study focuses on safety in the automotive industry and on one specific type of analysis. However, both the prompt patterns and decomposition techniques presented in this paper can be expanded and adapted to other methods of analysis and fields, such as quality or cybersecurity, including Threat Analysis \& Risk Assessment.
The suggested approach outlined in this study can be applied to various automotive functions. We chose an autonomous driving capability because it reduces the risk of data leakage into the AI's training data and presents more difficulties due to its novel challenges.



\section{Conclusions and Further Work} \label{sec:conclusion}

Analysing highly complex scenarios that an AD function can operate in requires vast knowledge to analyse the potential effect of each failure mode on the environment, which may not exist in legacy documentation. 
This is why LLMs, with their vast training data and ability to interpret further documents in natural language, might play a crucial role in automating tasks such as a HARA. 
The goal of this study was to design an LLM-based prototype capable of effectively supporting human engineers in specifying and specifying safety requirements in the context of complex automotive functions.

The findings from each engineering cycle contribute to the design of the prototype: Firstly, by identifying the limitations of using LLMs for specifying safety requirements for AD functions (\textbf{RQ1}), followed by addressing these limitations by breaking down HARA into tasks manageable for the LLM (\textbf{RQ2}) and refining the prompts to enhance the LLM's performance in generating safety requirements (\textbf{RQ3}). Independent safety experts finally evaluated the LLM-generated HARA in a stand-alone way and compared to a human-engineered baseline. Therefore, the tool not only demonstrates efficiency, completing tasks in one day compared to the months required by human effort, but its output also proves to be effective.



There are some other safety activities, to which the same concept can be applied to and to improve the efficiency. Using the tool to review and finding inconsistencies between requirements was also identified to be investigated in future studies. 
As one of the participants stated: ``there is a lot of room of using AI tools in system design and safety documents, this would be very good initiation to enter this journey to help experts increase productivity.''

Further work could also focus on providing a set of requirements in different abstraction levels and automated code generation generate the code. We need to consider that a requirement can also be considered as a prompt for another LLM, which may requires in tuning the process and rules to be used in prompt engineering to create a pipeline to create a deployable artifacts eventually. 

We expect the challenges encountered throughout the design cycles, as well as the safety experts' views, inspire further research on automating RE tasks within the automotive field. We also expect that this study contributes to highlight the need of human oversight, in this case VR and CR, when using LLMs for any safety-related activity.

\section*{Acknowledgments}
Thanks to Sweden's Innovation Agency (Vinnova) for funding (Diarienummer: 2021-02585), and WASP, for supporting this work.

\balance

\section*{Disclaimer}
The views and opinions expressed are those of the authors and do not necessarily reflect the official policy or position of Volvo Cars.
The proposed methods, prompts, pipeline or the results generated by this work are only used in this study and not used in any engineering of production related projects.

\end{document}